%% file: zz_workshop_papers/04-hierarchy_xai.tex
\documentclass[10pt,twocolumn,letterpaper]{article}

\usepackage{cvpr}      


\definecolor{cvprblue}{rgb}{0.21,0.49,0.74}
\usepackage[pagebackref,breaklinks,colorlinks,allcolors=cvprblue]{hyperref}

\usepackage{amsmath}
\DeclareMathOperator*{\argmin}{\arg\!\min}
\usepackage{booktabs}
\usepackage{multirow}

\def\paperID{****} 
\def\confName{CVPR}
\def\confYear{2025}

\title{Analyzing Hierarchical Structure in Vision Models with Sparse Autoencoders}

\author{
Matthew L. Olson$^{1,*}$, Musashi Hinck$^{1,*}$, Neale Ratzlaff$^1$, Changbai Li$^2$, \\
Phillip Howard$^1$, Vasudev Lal$^1$, Shao-Yen Tseng$^1$ \\
$^1$Intel Labs, Santa Clara, CA, USA \quad $^2$Oregon State University, Corvallis, OR, USA \\
$^1${\tt\small {\{matthew.lyle.olson, musashi.hinck, neale.ratzlaff,phillip.r.howard,}}\\
{\tt\small { vasudev.lal, shao-yen.tseng\}@intel.com}, 
$^2$\texttt{lc@oregonstate.edu}} \\
\small {$^*$Equal contributions}
}

\begin{document}


\maketitle

\setlength{\abovedisplayskip}{3pt} \setlength{\abovedisplayshortskip}{3pt}

\begin{abstract}
The ImageNet hierarchy provides a structured taxonomy of object categories, offering a valuable lens through which to analyze the representations learned by deep vision models. In this work, we conduct a comprehensive analysis of how vision models encode the ImageNet hierarchy, leveraging Sparse Autoencoders (SAEs) to probe their internal representations. SAEs have been widely used as an explanation tool for large language models (LLMs), where they enable the discovery of semantically meaningful features. Here, we extend their use to vision models to investigate whether learned representations align with the ontological structure defined by the ImageNet taxonomy. Our results show that SAEs uncover hierarchical relationships in model activations, revealing an implicit encoding of taxonomic structure. We analyze the consistency of these representations across different layers of the popular vision foundation model DINOv2 and provide insights into how deep vision models internalize hierarchical category information by increasing information in the class token through each layer. Our study establishes a framework for systematic hierarchical analysis of vision model representations and highlights the potential of SAEs as a tool for probing semantic structure in deep networks.
\end{abstract}

\section{Introduction}
The hierarchical structure of object categories plays an important role in human perception and cognition, influencing how we classify, recognize, and relate visual entities~\cite{peelen2017category}. In the context of computer vision, hierarchical taxonomies, such as those defined in ImageNet~\cite{russakovsky2015imagenet}, provide a structured organization of categories that can serve as a useful reference for analyzing how deep neural networks represent visual concepts. Understanding whether and how vision models internalize such hierarchical relationships is an open question in model explainability.

Sparse Autoencoders (SAEs) have emerged as a powerful tool for probing high-dimensional representations, particularly in the study of large language models (LLMs)~\cite{cunningham2023sparse,rajamanoharan2024gated_sae,gao2024scaling}. By enforcing sparsity in a learned bottleneck layer, SAEs extract disentangled features that correspond to meaningful latent factors in model activations. In this work, we apply SAEs to the analysis of vision models, using them to investigate whether model-internal representations reflect the hierarchical structure of ImageNet. Specifically, we aim to determine whether learned features naturally align with the taxonomy and how this alignment varies across different architectures and training methods.

We perform a detailed case study evaluating the of hierarchical encoding in the popular unsupervised vision foundation model DINOv2~\cite{oquab2023dinov2}, and leverage SAEs to extract and analyze sparse feature representations. Our study addresses the following key questions:
\begin{itemize}
\item Do the internal representations of vision models align with the ImageNet hierarchy, and can SAEs reveal this structure?
\item How consistent is the hierarchical structure across different layers of the model?
\item Can SAE-derived representations quantify the degree to which a model respects taxonomic relationships?
\end{itemize}

To answer these questions, we fit SAEs to every intermediate activation of DINOv2 model with respect to the ImageNet dataset and examine how the learned sparse features correspond to hierarchical category relationships. Our results indicate, 1) that DINOv2 does not encode information into the class token in early layers, and 2) SAEs recover a meaningful decomposition of representations in later layers, with extracted features reflecting ImageNet’s taxonomic structure. 

Our contributions are as follows:
\begin{enumerate}
\item We introduce SAEs as a tool for analyzing hierarchical structure in vision models.
\item We establish a framework and metrics for quantifying hierarchical consistency.
\item We present an empirical study on the extent to which deep vision models encode the ImageNet taxonomy, revealing hierarchical representation in later layers.
\end{enumerate}

\section{Background}
\paragraph{Sparse Autoencoders}

While there are many type of SAEs recently proposed~\cite{cunningham2023sparse, gao2024scaling, rajamanoharan2024gated_sae}, in this work we focus on the simplest model: the  ReLU SAE.

The ReLU SAE follows a straight forward setup. Given an input vector $x \in \mathbb{R}^n$ from the model representation space, the encoder and decoder are defined as:
\begin{align}
    z &= \text{ReLU}(W_{\text{enc}}x  + b_{\text{enc}}) \\
    \hat{x} &= W_{\text{dec}}z + b_{\text{dec}}
\end{align}
where $W_{\text{enc}} \in \mathbb{R}^{n \times d}$, $b_{\text{enc}} \in \mathbb{R}^n$, $W_{\text{dec}} \in \mathbb{R}^{d \times n}$, and $b_{\text{dec}} \in \mathbb{R}^d$. The loss function consists of reconstruction error and an $L_1$ sparsity penalty:
\begin{equation}
    L = \|x - \hat{x}\|^2_2 + \lambda \|z\|_1.
\end{equation}

Using both a large hidden size $d$ and an $L_1$ sparsity penalty, SAEs learn more monosemantic representations~\cite{olah2020zoom}, where each neuron encodes a single concept.

\paragraph{Related Work}
\citet{bilal2017convolutional} developed an interface to probe if convolutional encoders learned the ImageNet class hierarchy. Prior work has even attempted to train vision models directly on class hierarchies \citep{xia2023hgclip}. SAE's have been used to probe for hierarchical features in language models \citep{li2024geometry}, but have not been explored for visual hierarchies.
However, SAEs have been shown to localize salient features when trained on vision models \citet{fry2024saemultimodal}. SAEs have also been used to steer diffusion models \citet{daujotas2024interpreting} towards learned attributes. 
Finally, \citet{stevens2025sparse} studied SAEs trained on the patch embeddings of image encoder models.
In this work, we are the first to apply SAEs to analyze the ontological fidelity of the learned features of a vision foundation model with respect to the ImageNet class hierarchy. 

\section{Methods}

Image encoders, also known as visual foundation models are typically trained with self-supervised objective, and used to extract dense representations of visual inputs for other downstream tasks such as zero-shot image classification, object detection and captioning.
In our experiments we study the state-of-the-art unsupervised model DINOv2, which features a broad set of general visual capabilities without fine-tuning.  

In this work, we test whether SAEs capture \textit{ontological features} in vision encoders by identifying SAE heads that map to higher-order concepts. The ImageNet classes are drawn from the WordNet ontology, which relates these classes as a hierarchical tree of \textit{synsets}. We identify SAE heads that activate on groups of ImageNet classes that belong to the same higher-level WordNet concept. 

In this section, we first describe how we design metrics that capture the hierarchical learning via SAEs, then we discuss the results of our experiments showing SAEs capture complex semantic structures within image encoder models.

\subsection{Datasets}

ImageNet \cite{imagenet_dataset} is a large-scale visual dataset with over 1 million labeled images spanning thousands of object categories, widely used for training and evaluating image classification models. 

We leverage the hierarchical structure of the ImageNet classes.
The ImageNet-1k dataset contains $1000$ classes, which are synsets in the WordNet ontology.
The parents of a synset are \textit{hypernyms}, and its children are \textit{hyponyms}.
For example, \textit{Pembroke Corgi} is an ImageNet class, which is an hyponym of \textit{Corgi}, which is in turn a hyponym of \textit{Dog}. \textit{Dog} is a hypernym of \textit{Corgi} and \textit{Pembroke Corgi}, and \textit{Corgi} is a hypernym of \textit{Pembroke Corgi}.

\subsection{Hierarchical Metrics}
We are interested not only in SAE heads that activate on the $1000$ leaf-level classes, but also the extent to which SAE heads that activate on multiple classes are capturing higher levels in the ImageNet concept hierarchy. To measure this, we construct two metrics, which we call \textit{Lowest Common Hypernym (LCH) Height} and \textit{Ontological Coverage}.

Let $\Omega$ be the set of leaf ImageNet classes ($|\Omega|=1000$ for ImageNet-1k), and let  $\mathcal{S}$ be the set of all WordNet synsets that occur as ancestors (including self) of any leaf in $\Omega$. Thus $\Omega \subset \mathcal{S}$, but $\mathcal{S}$ is not the set of all WordNet synsets. For each synset $h \in \mathcal{S}$, we denote its leaf set as:
\begin{equation}
    L(h) = \{\omega \in \Omega: \text{there is a hypernym path }\omega\text{ to }h\}
\end{equation}

Thus, given the synset $h$, $L(h)$ is the set of all ImageNet leaf classes that are descendants (hyponyms) of $h$. Note also that for all $\omega \in \Omega$, $L({\omega})=\{\emptyset\}$.

We denote the set of classes an SAE head activates on as $C_k \subseteq \Omega$.
For a given set $C_k$, the \textit{lowest common hypernym} $h_k$ is the synset with the smallest subtree that contains all elements of $C_k$. This is analogous to \textit{lowest common ancestor}.
\begin{equation}
    h_k = \text{LCH}(C_k) = \argmin_{h \in \mathcal{S}:\;C_k \subseteq L_h}|L({h_k})|
\end{equation}

\vspace{-1.0em}
\paragraph{LCH Height} of $C_k$ is calculated the average height of $h_k$. This is equivalent to the average path distance between the elements of $C_k$ and $h_k$:
\begin{equation}
    \text{LCH Height}(C_k) = \frac{1}{|C_k|} \sum_{\omega \in C_k}\text{dist}(\omega, h_k)
\end{equation}

\vspace{-1.0em}
\paragraph{Ontological Coverage} is calculated as the proportion of elements in $C_k$ that are in $S_{h_k}$:
\begin{equation}
    \text{Coverage}(C_k) = \frac{|C_k|}{|L(h_k)|}
\end{equation}

These two metrics measure how well an SAE head captures a higher-order class in the ImageNet hierarchy. LCH height indicates the relevant level of abstraction that an SAE head may be capturing, and ontological coverage indicates how well it captures a higher-order concept.
There are some limitations to this metric: SAE heads activating on a single ImageNet class will have a coverage of $1$, and the use of the LCH as the relevant set may overly penalize heads that are largely coherent except for a single element. We thus consider the two metrics in tandem.

\subsection{Relevancy Maps}
We assess the spatial alignment of SAE features using relevancy maps \cite{chefer2021generic}, which highlight input regions contributing to the model’s output. Unlike traditional attention visualization, this method assigns local relevancy scores by computing gradient-weighted attention:

\begin{equation}
\bar{A} = \mathbb{E}_h\left((\nabla A \odot A)^+\right)
\end{equation}

Relevancy propagates across layers using:

\begin{equation}
R_{i} = R_{i-1} + \bar{A_{i}} \cdot R_{i-1}
\end{equation}

where \(R_i\) is initialized as an identity matrix. The final scores are row-normalized, excluding the identity contribution. See \citet{chefer2021generic} for more details.

\subsection{SAE training metrics}
\paragraph{MSE Reconstruction Error.} The reconstruction quality is evaluated using the mean squared error (MSE) between the input $x$ and the reconstructed output $\hat{x}$. We define MSE as: $\frac{1}{n} \|x - \hat{x}\|_2^2$

\vspace{-1.0em}
\paragraph{L1 Sparsity.} The L1 norm of the latent representation $z$ quantifies the level of sparsity after training:
$\|z\|_1 = \sum_{i=1}^{n} |z_i| $

\vspace{-1.0em}
\paragraph{L0 Activation Count} The L0 norm measures the number of active (nonzero) latent units: $|z\|_0 = \sum_{i=1}^{n} \mathbf{1}(z_i \neq 0)$
where $\mathbf{1}(\cdot)$ is an indicator function. This metric directly quantifies the sparsity level by counting active units.

\subsection{Implementation Details}

The input to the SAEs are the class embedding output by the base image encoder at a selected layer. We train a total of 40 SAEs, 1 for each layer of DINOv2.
All SAEs are trained for three epochs and minibatch of size $64$, with an Adam\cite{kingma2014adam} optimizer using a $1e^{-4}$ learning rate with  5\% linear warm-up and 20\% linear decay. We also use a 5\% $\lambda$ warm-up to minimize dead neurons. For all experiments, we use $\lambda=10$ as a trade-off between reconstruction quality, while ensuring sparsity. Additionally, we use a hidden expansion factor of $8$, resulting in an SAE hidden size of $12,288$.
We use the SAELens library~\cite{bloom2024saetrainingcodebase} for our training. Lastly, images are resized to $224\times224$.

We also train 40 linear probe models to measure the classification accuracy at each layer. These linear models are independently placed at each layer of DINOv2. They use identical training parameters as the SAEs where relevent.

\section{Results}

\begin{figure}[tb]
    \includegraphics[width=\linewidth]{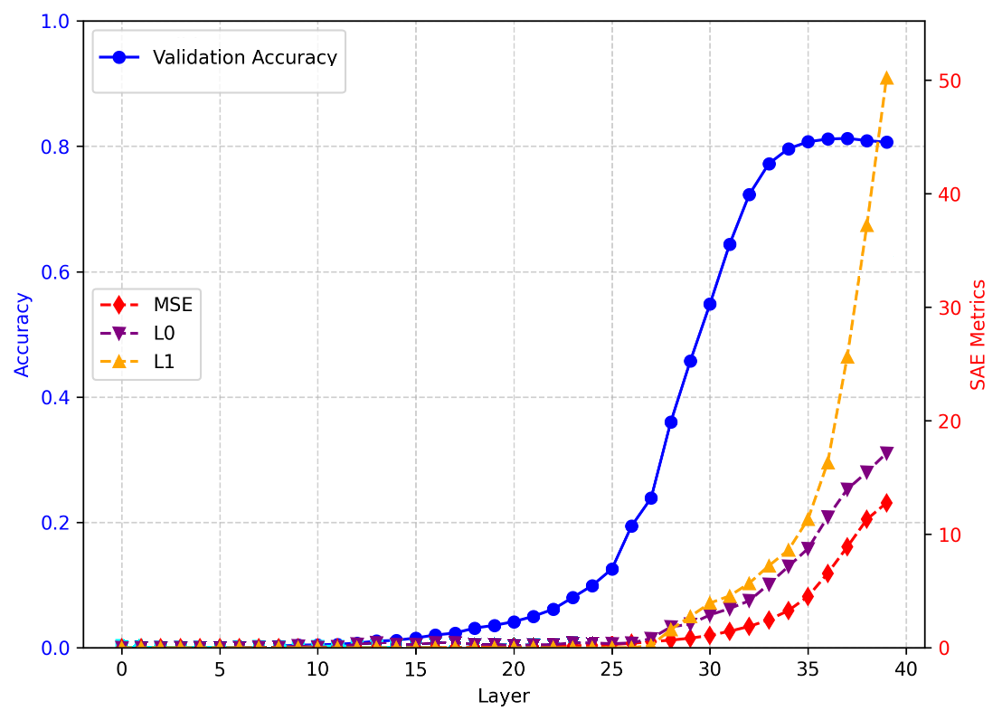}
    \caption{\textbf{Results of training a ReLU SAE (or linear probe) on every layer of DINOv2's class token on ImageNet.} We find the surprising result that the early layers in this model are non-informative: the representations are incredibly easy to auto-encode (right y-axis), require very few activations from an SAE (right y-axis), and are not usable for fitting a classification model (left y-axis). }
    \label{fig:dino_by_layer}
\end{figure}

In figure \ref{fig:dino_by_layer} we present the results of our first experimental analysis. We find the interesting result that early layers of DINOv2's class representation contain no information. As models are fit later in the layers, the better they do at classification and the worse they do in SAE metrics. This implies the representation at each layer gains more information as the model processes the input image tokens. The results suggest SAEs can be used as a surrogate to identify information in a given model's token representations without the need for labels simply by measuring the unsupervised SAE metrics.

\subsection{Ontological Features}

\begin{figure}
    \centering
    \includegraphics[width=1.0\linewidth]{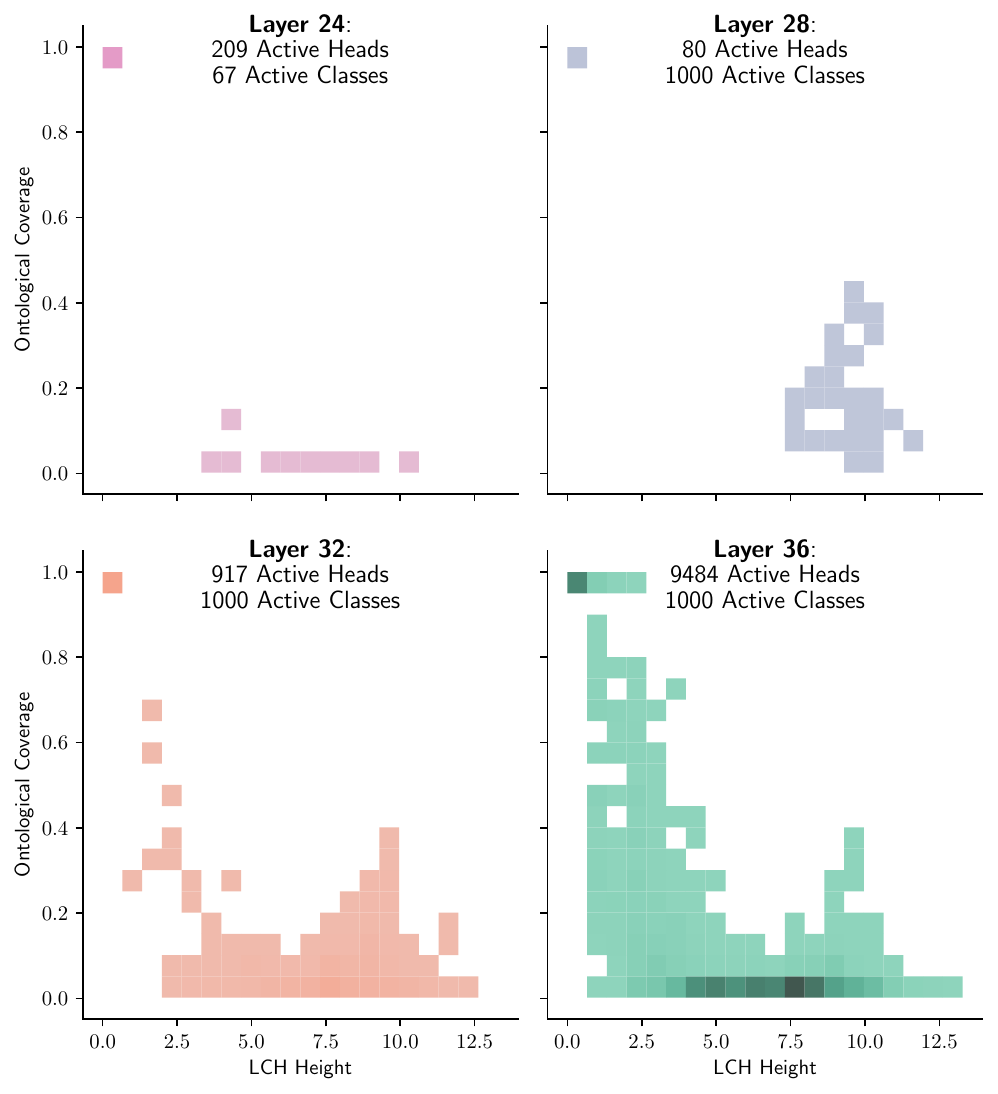}
    \caption{
    \textbf{Distribution of LCH Height vs Ontological Coverage for SAE Heads at Layer 24, 28, 32 and 36 of DINOv2.}
    For each layer, we plot the distribution of LCH height and ontological coverage of the SAE heads. Darker indicates higher bin density. Not only does the vision model capture hierarchical concepts in its output, but also show signs of enhancing hierarchical features through out its processing layer-by-layer.
    }
    \label{fig:lch_coverage}
\end{figure}

Figure \ref{fig:lch_coverage} shows the results on ImageNet validation set's distribution of coverage and LCH height for SAE heads from layer 24, 28, 32, and 36 of DINOv2 -- given how there is little SAE head activation in earlier layers the model.

SAE heads at layer 24 mostly activate either on a single class (top-left corner of the top-left subplot) or on many dissimilar classes (strip along the bottom). One interesting exception is head 657, which activates on 7 bird species and has a coverage of $0.119$.

Later layers have increasing numbers of SAE heads with high ontological coverage. In particular, layer 36 has 90 multi-class SAE heads with ontological coverage of $1.0$, capturing higher-order groups of things such as \textit{elasmobranch} (sharks and rays), \textit{whales}, \textit{woodwind instruments} and \textit{warships}. 
Our findings show that foundation models like DINOv2 capture hierarchical concepts, with SAEs serving as a powerful tool for elucidating these results.

\begin{figure}
    \centering
    \includegraphics[width=0.9\linewidth]{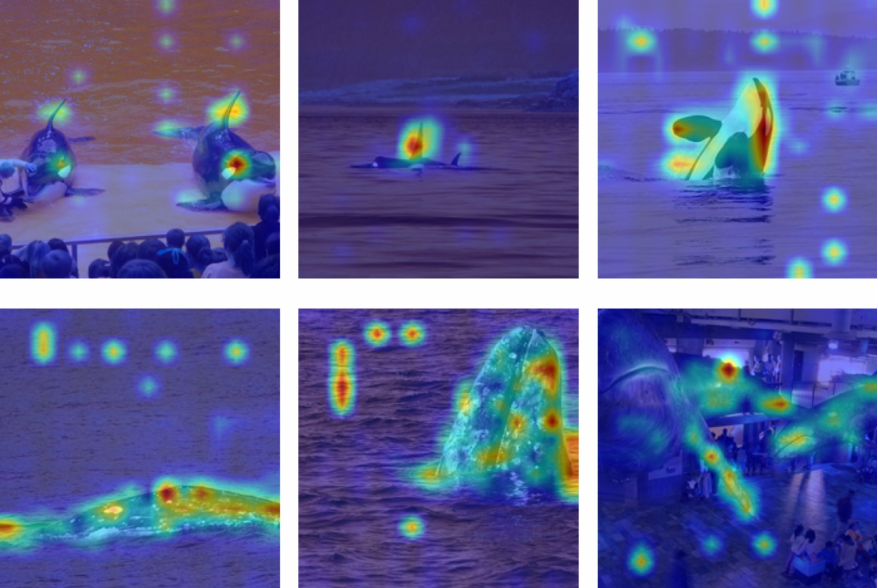}
    \caption{
    Relevancy maps of the hierarchical SAE head at DINOv2 Layer 36 activating on images of whales. These relevancy maps show the model highly activating on the hierarchical concept of both Orcas and Grey Whales, which show DINOv2's ability to focus on highly meaningful parts of an image.
    }
    \label{fig:xai_rm}
\end{figure}

\subsection{Hierarchical Relevancy Maps}
We show an example of the relevancy map, as shown in Figure \ref{fig:xai_rm}
Given an image $I$, we generate feature-wise heatmaps highlighting important regions responsible for the activation of each sparse feature, providing insight into the grounding of interpretable features. These results point towards the vision model's ability to encode semantically meaningful and hierarchical concepts and how SAEs can extract such information from the base model.

\section{Discussion}
We examined how deep vision models encode hierarchical relationships in the ImageNet taxonomy. Using Sparse Autoencoders (SAEs) as a probe, we found that taxonomic structures are partially reflected in model representations. SAEs extract disentangled features aligned with hierarchy, with early layers showing stronger alignment. These results highlight SAEs as a useful tool for structured explanations of features in vision models.

\paragraph{Limitations}
While our findings provide valuable insights, several limitations must be acknowledged. First, our study is limited to a single vision foundation model, DINOv2, and may not generalize to all architectures, particularly those with different training paradigms or inductive biases. Second, our analysis primarily focuses on the ImageNet hierarchy, which, while widely used, is not necessarily the most comprehensive or universally applicable taxonomy for visual concepts. Third, the reliance on SAEs introduces its own interpretability challenges, such as the potential for feature redundancy or artifacts introduced by the sparsity constraint. Finally, our hierarchical metrics, while informative, may not fully capture all nuances of taxonomic representation within vision models, necessitating further refinement and alternative evaluation strategies.

\paragraph{Future Work}
Future research could extend our analysis to diverse models and taxonomies, including those from human perception studies. Advanced XAI methods, such as causal interventions, may further clarify hierarchical encoding. Finally, SAEs could enable applications in hierarchical classification and concept-based model editing.

{\small
\bibliographystyle{ieeenat_fullname}
\bibliography{egbib}
}

\end{document}


%% file: zz_workshop_papers/04-hierarchy_xai.bbl
\begin{thebibliography}{17}
\providecommand{\natexlab}[1]{#1}
\providecommand{\url}[1]{\texttt{#1}}
\expandafter\ifx\csname urlstyle\endcsname\relax
  \providecommand{\doi}[1]{doi: #1}\else
  \providecommand{\doi}{doi: \begingroup \urlstyle{rm}\Url}\fi

\bibitem[Bilal et~al.(2017)Bilal, Jourabloo, Ye, Liu, and Ren]{bilal2017convolutional}
Alsallakh Bilal, Amin Jourabloo, Mao Ye, Xiaoming Liu, and Liu Ren.
\newblock Do convolutional neural networks learn class hierarchy?
\newblock \emph{IEEE transactions on visualization and computer graphics}, 24\penalty0 (1):\penalty0 152--162, 2017.

\bibitem[Chefer et~al.(2021)Chefer, Gur, and Wolf]{chefer2021generic}
Hila Chefer, Shir Gur, and Lior Wolf.
\newblock Generic attention-model explainability for interpreting bi-modal and encoder-decoder transformers.
\newblock In \emph{Proceedings of the IEEE/CVF international conference on computer vision}, pages 397--406, 2021.

\bibitem[Cunningham et~al.(2023)Cunningham, Ewart, Riggs, Huben, and Sharkey]{cunningham2023sparse}
Hoagy Cunningham, Aidan Ewart, Logan Riggs, Robert Huben, and Lee Sharkey.
\newblock Sparse autoencoders find highly interpretable features in language models.
\newblock \emph{arXiv preprint arXiv:2309.08600}, 2023.

\bibitem[Daujotas(2024)]{daujotas2024interpreting}
G. Daujotas.
\newblock Interpreting and steering features in images.
\newblock \url{https://www.lesswrong.com/posts/Quqekpvx8BGMMcaem/interpreting-and-steering-features-in-images}, 2024.
\newblock Accessed: 2025-03-07.

\bibitem[Deng et~al.(2009)Deng, Dong, Socher, Li, Li, and Fei-Fei]{imagenet_dataset}
Jia Deng, Wei Dong, Richard Socher, Li-Jia Li, Kai Li, and Li Fei-Fei.
\newblock Imagenet: A large-scale hierarchical image database.
\newblock In \emph{2009 IEEE conference on computer vision and pattern recognition}, pages 248--255. Ieee, 2009.

\bibitem[Fry(2024)]{fry2024saemultimodal}
Hugo Fry.
\newblock Towards multimodal interpretability: Learning sparse interpretable features in vision transformers.
\newblock \url{https://www.lesswrong.com/posts/bCtbuWraqYTDtuARg/towards-multimodal-interpretability-learning-sparse-2}, 2024.

\bibitem[Gao et~al.(2024)Gao, la~Tour, Tillman, Goh, Troll, Radford, Sutskever, Leike, and Wu]{gao2024scaling}
Leo Gao, Tom~Dupr{\'e} la Tour, Henk Tillman, Gabriel Goh, Rajan Troll, Alec Radford, Ilya Sutskever, Jan Leike, and Jeffrey Wu.
\newblock Scaling and evaluating sparse autoencoders.
\newblock \emph{arXiv preprint arXiv:2406.04093}, 2024.

\bibitem[Joseph~Bloom and Chanin(2024)]{bloom2024saetrainingcodebase}
Curt~Tigges Joseph~Bloom and David Chanin.
\newblock Saelens.
\newblock \url{https://github.com/jbloomAus/SAELens}, 2024.

\bibitem[Kingma and Ba(2014)]{kingma2014adam}
Diederik~P Kingma and Jimmy Ba.
\newblock Adam: A method for stochastic optimization.
\newblock \emph{arXiv preprint arXiv:1412.6980}, 2014.

\bibitem[Li et~al.(2024)Li, Michaud, Baek, Engels, Sun, and Tegmark]{li2024geometry}
Yuxiao Li, Eric~J Michaud, David~D Baek, Joshua Engels, Xiaoqing Sun, and Max Tegmark.
\newblock The geometry of concepts: Sparse autoencoder feature structure.
\newblock \emph{arXiv preprint arXiv:2410.19750}, 2024.

\bibitem[Olah et~al.(2020)Olah, Cammarata, Schubert, Goh, Petrov, and Carter]{olah2020zoom}
Chris Olah, Nick Cammarata, Ludwig Schubert, Gabriel Goh, Michael Petrov, and Shan Carter.
\newblock Zoom in: An introduction to circuits.
\newblock \emph{Distill}, 5\penalty0 (3):\penalty0 e00024--001, 2020.

\bibitem[Oquab et~al.(2023)Oquab, Darcet, Moutakanni, Vo, Szafraniec, Khalidov, Fernandez, Haziza, Massa, El-Nouby, et~al.]{oquab2023dinov2}
Maxime Oquab, Timoth{\'e}e Darcet, Th{\'e}o Moutakanni, Huy Vo, Marc Szafraniec, Vasil Khalidov, Pierre Fernandez, Daniel Haziza, Francisco Massa, Alaaeldin El-Nouby, et~al.
\newblock Dinov2: Learning robust visual features without supervision.
\newblock \emph{arXiv preprint arXiv:2304.07193}, 2023.

\bibitem[Peelen and Downing(2017)]{peelen2017category}
Marius~V Peelen and Paul~E Downing.
\newblock Category selectivity in human visual cortex: Beyond visual object recognition.
\newblock \emph{Neuropsychologia}, 105:\penalty0 177--183, 2017.

\bibitem[Rajamanoharan et~al.(2024)Rajamanoharan, Conmy, Smith, Lieberum, Varma, Kram{\'a}r, Shah, and Nanda]{rajamanoharan2024gated_sae}
Senthooran Rajamanoharan, Arthur Conmy, Lewis Smith, Tom Lieberum, Vikrant Varma, J{\'a}nos Kram{\'a}r, Rohin Shah, and Neel Nanda.
\newblock Improving dictionary learning with gated sparse autoencoders.
\newblock \emph{arXiv preprint arXiv:2404.16014}, 2024.

\bibitem[Russakovsky et~al.(2015)Russakovsky, Deng, Su, Krause, Satheesh, Ma, Huang, Karpathy, Khosla, Bernstein, Berg, and Fei-Fei]{russakovsky2015imagenet}
Olga Russakovsky, Jia Deng, Hao Su, Jonathan Krause, Sanjeev Satheesh, Sean Ma, Zhiheng Huang, Andrej Karpathy, Aditya Khosla, Michael Bernstein, Alexander~C. Berg, and Li Fei-Fei.
\newblock Imagenet large scale visual recognition challenge, 2015.

\bibitem[Stevens et~al.(2025)Stevens, Chao, Berger-Wolf, and Su]{stevens2025sparse}
Samuel Stevens, Wei-Lun Chao, Tanya Berger-Wolf, and Yu Su.
\newblock Sparse autoencoders for scientifically rigorous interpretation of vision models.
\newblock \emph{arXiv preprint arXiv:2502.06755}, 2025.

\bibitem[Xia et~al.(2023)Xia, Yu, Hu, Ju, Wang, Duan, and Ge]{xia2023hgclip}
Peng Xia, Xingtong Yu, Ming Hu, Lie Ju, Zhiyong Wang, Peibo Duan, and Zongyuan Ge.
\newblock Hgclip: exploring vision-language models with graph representations for hierarchical understanding.
\newblock \emph{arXiv preprint arXiv:2311.14064}, 2023.

\end{thebibliography}
